# Simulating Field Experiments with Large Language Models

Yaoyu Chen

University of Illinois at Chicago

ychen563@uic.edu

Yuheng Hu

University of Illinois at Chicago

yuhenghu@uic.edu

Yingda Lu

University of Illinois at Chicago

yingdalu@uic.edu

## Abstract

Prevailing large language models (LLMs) are capable of human responses simulation through its unprecedented content generation and reasoning abilities. However, it is not clear whether and how to leverage LLMs to simulate field experiments. In this paper, we propose and evaluate two prompting strategies: the observer mode that allows a direct prediction on main conclusions and the participant mode that simulates distributions of participants' responses. Using this approach, we examine fifteen well-cited field experimental papers published in INFORMS and MISQ, finding encouraging alignments between simulated experimental results and the actual results in certain scenarios. We further identify topics of which LLMs underperform, including gender difference and social norms related research. Additionally, the automatic and standardized workflow proposed in this paper enables the possibility of a large-scale screening of more papers with field experiments.

This paper pioneers the utilization of large language models (LLMs) for simulating field experiments, presenting a significant extension to previous work which focused solely on lab environments. By introducing two novel prompting strategies—observer and participant modes—we demonstrate the ability of LLMs to both predict outcomes and replicate participant responses within complex field settings. Our findings indicate a promising alignment with actual experimental results in certain scenarios, achieving a stimulation accuracy of 66% in observer mode. This study expands the scope of potential applications for LLMs and illustrates their utility in assisting researchers prior to engaging in expensive field experiments. Moreover, it sheds light on the boundaries of LLMs when used in simulating field experiments, serving as a cautionary note for researchers considering the integration of LLMs into their experimental toolkit.

## Introduction

Field experiments allow researchers to manipulate variables of interest in a real-world setting, facilitating the establishment of causal relationships between interventions and outcomes, a key advantage over more controlled but less realistic lab studies. Moreover, unlike lab studies that take place in controlled environments, field experiments are conducted in natural settings, which significantly enhances the relevance and applicability of the findings to actual economic phenomena, policies, and decision-making, making the results from the experiments more generalizable to real-world contexts. Given these advantages, field experiments have become increasingly popular over the years in both academia and industry (Dennis & Valacich, 2001).

The rapid advancements in Generative AI, particularly with the release of sophisticated Large Language Models (LLMs), have led scholars to explore the potential of these models for simulating human responses and behaviors, thus gaining novel insights in fields such as psychology (Aher et al., 2023), sociology (Manning et al., 2024), and economics (Horton, 2023). Specifically, these simulations involve LLMs as proxy participants in lab experiments, where the models are tasked with emulating human responses. Since field experiments are considered methodologically complementary to lab experiments, the feasibility of field experimental simulations is also worth exploring.

The paper presents a framework powered by GPT-4 that conducts experimental simulation in two modes, namely, observer mode and participant mode. Specifically, the framework takes field experimental settings in two alternatives as an input: either user-provided inputs or extracted data from existing papers. Whereas the former approach allows users to input their desirable settings to execute novel field experimental simulations, the latter way extracts settings of a field experiment from existing paper by Elicit.org, a LLM specialized in comprehending academic texts. Once the key settings are obtained through either method, the framework proceeds to simulate the field experiment in both observer and participant modes. In observer mode, we prompt the LLM to observe a target field experiment, gathering information about its configurations and settings and predicting the main conclusions. In participant mode, we prompt the LLM with different profiles, and the LLM then role-plays a participant in the experiment, receiving instructions and treatments from organizers and answering questions as requested. In other words, observer mode directly predicts the main conclusions, while participant mode generates simulated data distributions that could reflect the main conclusions.

We employ these two strategies on GPT-4 to simulate 15 field experiments adopted from existing marketing and information system literature. Our results show that observer mode is quite promising, successfully replicating over 66% of field experiments. By contrast, participant mode of GPT-4 achieved only 47.9%. Moreover, we find that results are highly skewed because of topic sensitivity. In particular, the LLM performs poorly in field experiments concerning gender differences, popularity information, humanizing customer service chatbots, and reciprocity.

## Prior Literature

Driven by their unprecedented reasoning, inference, and content-generation abilities, Large Language Models (LLMs) have found widespread application across both industry and academia. Among the various LLMs, the GPT (Generative Pre-trained Transformer) series by OpenAI stands out as one of the most prominent and widely used models.

Existing research suggests that LLMs can simulate how individuals behave in various economics or behavioral economics studies, particularly using LLMs to simulate participants in lab experiments (Aher et al., 2023; Horton, 2023; Leng & Yuan, 2024; Manning et al., 2024). Aher et al. (2023) and Horton (2023) follow a two-step process, which includes profiling agents and prompting experimental backstories, Leng and Yuan (2024) advance the framework by an interaction mechanism between agents to allow a more complex experimental setting. Recently, Manning et al. (2024) propose an automated framework that includes the aforementioned features for lab experimental simulation of scenarios regarding between-human interactions.

Notably, all existing LLM-powered experimental simulations are conducted for lab experiments. Lab experiments, however, are necessarily limited in relevance for predicting field behaviors (Harrison & List, 2004). By contrast, natural field experiments offer distinct advantages, such as improved econometric robustness and more effective control over participant recruitment. Promoting the adoption of behavioral field experiments has been a general shift, as evidenced by IS journals like Management Science (Gneezy, 2017). Moreover, extant studies merely propose prompting strategies and require the design of specific prompts for each laboratory experimental simulation (Aher et al., 2023; Horton, 2023; Leng & Yuan, 2024). By contrast, our novel framework, comprising observer and participant mode, introduces a unified approach to systematically examine the LLM's emergent abilities in simulating field experiments.

**Method**

Our process involves three steps, as illustrated in Figure 1. First, we select the top 40 most cited papers from the area of marketing and information systems in the INFORMS portal that contain "field experiment" in either the titles or abstracts, published between 2018 to 2024. We then manually examine these papers to ensure they are relevant to field experiments. As a result, 15 of these papers are qualified, as shown in Table 1.

Next, we need to extract information from these papers that characterize the field experiments. We rely on Elicit.org, a fine-tuned LLM trained on academic papers, to extract dependent variables, theoretical framework, duration, intervention, intervention effects, population characteristics, and main findings using. We also manually examine the extracted information to ensure its validity.

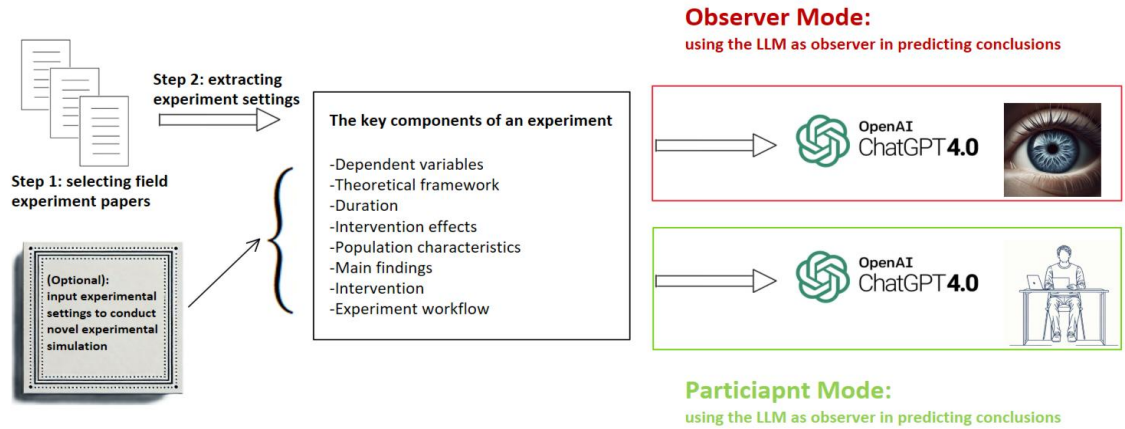

**Figure 1: The Workflow**

Note that OpenAI API (gpt-4-turbo-2024-04-09) is used through the entire process to ensure automation. pretrained Large Language Model specialized for the reading of academic papers (Elicit.org) is used in step 2 to extract setting information of field experiments in selected papers.

| No. | Paper |
|-----|-------|
| 1 | "Do women avoid salary negotiations? Evidence from a large-scale natural field experiment." (Leibbrandt et al., 2015), Management Science |
| 2 | "The impact of gender diversity on the performance of business teams: Evidence from a field experiment." (Hoogendoorn et al., 2013), Management Science |
| 3 | "Employee recognition and performance: A field experiment." (Bradler et al., 2016), Management Science |
| 4 | "How does popularity information affect choices? A field experiment." (Tucker & Zhang, 2011), Management Science |
| 5 | "Holding the hunger games hostage at the gym: An evaluation of temptation bundling." (Milkman et al., 2014), Marketing Science |
| 6 | "You've got mail: A randomized field experiment on tax evasion." (Bott et al., 2020), Management Science |
| 7 | "Reconfiguring for Agility: Examining the Performance Implications of Project Team Autonomy through an Organizational Policy Experiment." (Ramasubbu & Bardhan, 2021b), MIS Quarterly |
| 8 | "Why do stores drive online sales? Evidence of underlying mechanisms from a multichannel retailer." (Kumar et al., 2019), Information Systems Research |
| 9 | "Extrinsic versus intrinsic rewards for contributing reviews in an online platform." (Khern-am-nuai et al., 2018), Information Systems Research |
| 10 | "Estimating the impact of "humanizing" customer service chatbots." (Schanke et al., 2021), Information Systems Research |
| 11 | "Impact of Customer Compensation Strategies on Outcomes and the Mediating Role of Justice Perceptions: A Longitudinal Study of Target's Data Breach." (Hoehle et al., 2022), MIS Quarterly |
| 12 | "Effects of online recommendations on consumers' willingness to pay." (Adomavicius et al., 2018), Information Systems Research |
| 13 | "RECIPROCITY OR SELF-INTEREST? LEVERAGING DIGITAL SOCIAL CONNECTIONS FOR HEALTHY BEHAVIOR." (Liu et al., 2022), MIS Quarterly |
| 14 | "An empirical investigation of the antecedents and consequences of privacy uncertainty in the context of mobile apps." (Al-Natour et al., 2020), Information Systems Research |
| 15 | "Pictures that are worth a thousand donations: How emotions in project images drive the success of online charity fundraising campaigns? An image design perspective." (Hou et al., 2023), MIS Quarterly |

**Table 1: Selected Papers**

***Prompting for Observer Mode***

In observer mode, given key experimental settings as inputs, the LLM acts as an observer to predict the main conclusions of field experiments. We design a prompting strategy (Figure 2), which is relies on the information automatically extracted from step 2 (Figure 1). Specifically, key components from A to F and the question section at the bottom are populated using the extracted information, which obviously varies across different experiments. For example, for "C. Dependent Variables:" this line is directly available from Elicit.org, whereas "D. Participants Information:" this line is also completed with "Population characteristics" collected previously. Following these settings, a question section asks the LLM to predict the main conclusions of existing papers in a specified manner.

**Observer Prompting**

**User:**
We are doing a field experiment. The design of this experiment is provided below:
A. The general goal:
B. Treatment:
C. Dependent Variables:
D. Participants Information:
E. The Experiment Duration:
F. Additional Details about the Experiment Workflow:
You need to provide correct answers to questions related to the experiment above. Please answer the following questions in order with a good compliance with the instructions of each question.
Question No.{num}: which one will be the correct conclusion based on all experiment design information above? Possible conclusions are {option 1}, {option 2}, {option 3}. Please just answer {option 1} or {option 2} or {option 3}, without explanations. Be sure to format your response according to the template below: Question No.num: {the option you chose}

**AI:**

**Figure 2: Observer Prompting**

**The General Goal Generating Prompting**

**User:**
We are doing a field experiment. The design of this experiment is provided
I'm trying to summarize te general goal of a field experiment paper. The general goal follows the format: the impact of {the main treatment} on {the main dependent variable}. The title of this paper is {title}. The possible treatments are as follow: {treatments}. The dependent variables are as follow: {y}. Would you please help me complete the general goal using the format provided given all information above? Make sure that your response aligns with the format: the impact of {the main treatment} on {the main dependent variable}. Don't include any explanations.

**AI:**

**Figure 3: The General Goal Generating Prompting**

Sometimes not all information required in this template is directly available from the feature extraction by Elicit.org. For example, "A. The General Goal" is
7

unavailable from the extracted features. By including a series of key components such as treatments, dependent variables, and titles within the prompting template shown in Figure 3, the LLM can summarize the general goal of the corresponding paper according to the required format. Therefore, additional natural language processing is necessary to properly format the content to be added to the template, which is done by standalone GPT-4-turbo sessions.

In the question part of observer template, since the main findings from selected papers are extracted in step 2 of Figure 1, each question included in this part of the template corresponds to one of the main findings. As a result, the total number of questions in one observer prompting equals to the number of main findings of a paper. To test if the LLM can accurately predict the conclusions of field experiments, each finding is rephrased into two variants, reversed and non-related, by separate LLM sessions.

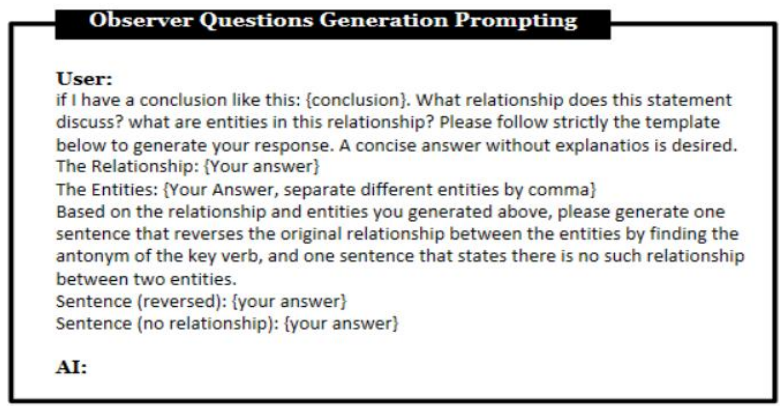

Specifically, as depicted in Figure 4, after a main finding is prompted to the LLM, it follows instructions to deconstruct the content and reorganize components into two variants of the original conclusion. The reversed variant means that the direction of the conclusion is inverted, while the non-related variant typically indicates that there is no causal relationship between the entities of interest. The original conclusion and the two variants are then included as a question in the observer template. The observer LLM is required to select one from the "three options" to demonstrate its ability to predict conclusions, inspired by Luo et al. (2024)'s prediction of neuroscience results by LLMs.

**Figure 4: The Observer Questions Generation Prompting**

### *Prompting for Participant Mode*

In this mode, the LLM acts as a participant in a field experiment, which receives instructions and makes choices exactly as a human participant would do in the actual field experiments. The method of instructing participants and the related experiment settings that participants should be aware of are extracted from the previous step, similar to those in observer mode. In the end, variants of instruction promptings from the second-person perspective are generated by the LLM using the generation prompting in Figure 5. Although each field experiment has a specific design, there are variants because different participants receive different treatments. Additionally, there is a need to profile participants since some studies focus on the impact of other factors, such as gender, resulting in more variants. For example, in simulating the first paper listed on Table 1, "Do women avoid salary negotiations? Evidence from a large-scale natural field experiment." (Leibbrandt et al., 2015), at least two variants of synthetic participants (male and female) need to be generated since the paper focuses on the gender difference of the treatment effect. After instruction prompts for participant mode are generated, they are concatenated with questions that participants are expected to answer in the field experiment, for example, "After the treatment, do you want to buy an additional unit of the product?"

## Results

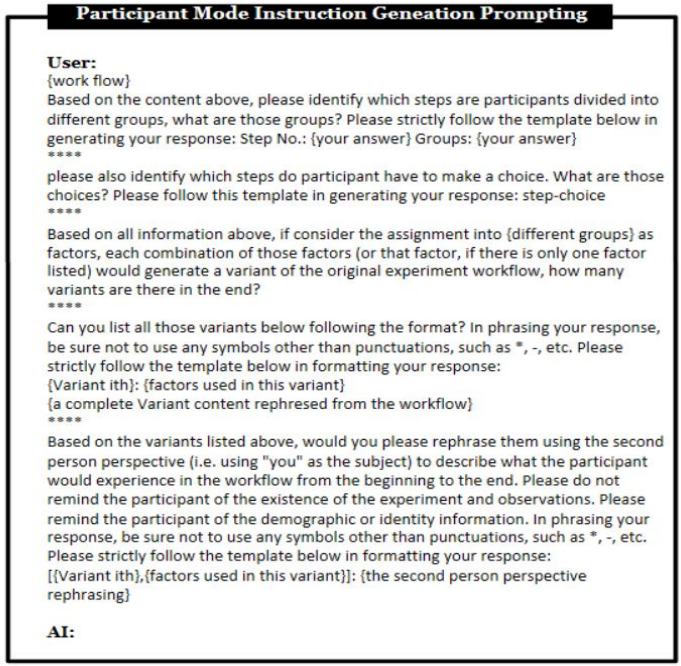

**Figure 5: Participant Mode Instruction Generation Prompting**

After completing the workflow in Figure 1, we apply our framework to each and every paper. With observer mode, each paper has one prompting generated according to the template in Figure 2. With the participant mode, each paper has at least two participant promptings since field experiments naturally divide participants into at least two groups. In different groups, participants experience different scenarios and receive different treatment. If the paper considers additional factors, such as gender, there could be more than two variants. For instance, assigning female and male participants into two groups could result in four variants of promptings.

We use OpenAI's GPT-4-turbo API to obtain simulation results and record the responses. API parameters, including seed and temperature, are set to default. In observer mode, we prompt each paper 30 times, with each response containing predictions on all hypotheses of that paper. For participant mode, since there are at least two variants of prompting, we repeat each variant five times. Each response is a direct answer from participants, such as a decision on whether to buy a product or not, representing a data point collected from an experiment.

### *Results of Observer Mode*

| No. | Paper | Observer Mode Results* | | | | | | |
|---|---|---|---|---|---|---|---|---|
| | | C1 | C2 | C3 | C4 | C5 | C6 | Avg (o) |
| 1 | "Do women avoid salary negotiations? Evidence from a large-scale natural field experiment." | 100% | 100% | 100% | 0% | | | 75% |
| 2 | "The impact of gender diversity on the performance of business teams: Evidence from a field experiment." | 0% | 0% | 80% | | | | 27% |
| 3 | "Employee recognition and performance: A field experiment." | 100% | 100% | 0% | | | | 67% |
| 4 | "How does popularity information affect choices? A field experiment." | 0% | 0% | 0% | | | | 0% |
| 5 | "Holding the hunger games hostage at the gym: An evaluation of temptation bundling." | 100% | 87% | 3% | 87% | | | 69% |
| 6 | "You've got mail: A randomized field experiment on tax evasion." | 0% | 100% | 60% | 100% | 100% | | 72% |
| 7 | "Reconfiguring for Agility: Examining the Performance Implications of Project Team Autonomy through an Organizational Policy Experiment." | 100% | 100% | | | | | 100% |
| 8 | "Why do stores drive online sales? Evidence of underlying mechanisms from a multichannel retailer." | 100% | 100% | 93% | 100% | | | 98% |
| 9 | "Extrinsic versus intrinsic rewards for contributing reviews in an online platform." | 83% | 0% | 0% | 100% | 100% | | 57% |
| 10 | "Estimating the impact of "humanizing" customer service chatbots." | 23% | 57% | 17% | | | | 32% |
| 11 | "Impact of Customer Compensation Strategies on Outcomes and the Mediating Role of Justice Perceptions: A Longitudinal Study of Target's Data Breach." | 100% | 100% | 100% | 100% | 100% | 100% | 100% |
| 12 | "Effects of online recommendations on consumers' willingness to pay." | 100% | 100% | 100% | | | | 100% |
| 13 | "RECIPROCITY OR SELF-INTEREST? LEVERAGING DIGITAL SOCIAL CONNECTIONS FOR HEALTHY BEHAVIOR." | 90% | 3% | 0% | | | | 31% |
| 14 | "An empirical investigation of the antecedents and consequences of privacy uncertainty in the context of mobile apps." | 100% | 100% | 10% | 100% | 100% | | 82% |
| 15 | "Pictures that are worth a thousand donations: How emotions in project images drive the success of online charity fundraising campaigns? An image design perspective." | 100% | 100% | 100% | 20% | 100% | 100% | 87% |

**\*c1 means "conclusion 1," which is the first conclusion of a paper. The framework automatically identifies the ordinal of conclusions.**

### Table 2: Observer Mode Results

The results of observer mode are shown in Table 2. Each paper contains between 2 to 6 main conclusions. We measure the accuracy of predictions by the LLM in the simulations using the following formula: $Accuracy\,(o) = N(predicted\ conclusion \equiv actual\ conclusion)\ /\ 30$, where the numerator indicates the number of correct predictions, and the denominator 30 indicates the number of times observer mode template (Figure 2) was prompted to the LLM. As a result, "1" means that the LLM could always choose the correct conclusions in all 30 trials. Conceivably, Avg (o) is the average of observer mode's accuracies of a paper: $Avg\,(o) = \frac{\sum_{1st\ conclusion}^{the\ last\ conclusion} accuracy\,(o)}{Number\ of\ conclusions}$.

After checking the key words of the 15 papers and comparing them with the results in Table 2, we find that the result is skewed and sensitive to topics. Whereas the LLM achieves high simulation accuracy in topics such as willingness-to-pay (the 12th and 14th papers) and privacy (11th and 14th papers), it performs poorly on the following topics: gender difference (the 2nd paper), popularity information (the 4th paper), humanizing customer service chatbots (the 10th paper), and reciprocity (the 13th paper). For the gender-related topic, existing research has found there is a substantial gender bias in earlier versions of GPT-4 (Zhao et al., 2024). Popularity information, humanizing bots, and reciprocity are all related to social norms. Even the most advanced LLMs cannot achieve a human-level understanding of social norms (Yuan et al., 2024). We argue that a lack of awareness of social pressures could possibly result in deviations of simulated experimental results from actual results in the topics of popularity information and reciprocity. This issue has the potential to improve as researchers are enhancing the awareness of social norms in LLMs.

### *Results of Participant Mode*

| No. | Paper | Participant Mode Results* | | | | | | |
|---|---|---|---|---|---|---|---|---|
| | | C1 | C2 | C3 | C4 | C5 | C6 | Avg (p) |
| 1 | "Do women avoid salary negotiations? Evidence from a large-scale natural field experiment." | 0 | 0 | 0 | 0 | | | 0% |
| 2 | "The impact of gender diversity on the performance of business teams: Evidence from a field experiment." | 0 | 0 | 0 | | | | 0% |
| 3 | "Employee recognition and performance: A field experiment." | 1 | 1 | x | | | | 100% |
| 4 | "How does popularity information affect choices? A field experiment." | 0 | 1 | | | | | 33% |
| 5 | "Holding the hunger games hostage at the gym: An evaluation of temptation bundling." | 0 | 0 | 0 | 0 | | | 0% |
| 6 | "You've got mail: A randomized field experiment on tax evasion." | 0 | 0 | 0 | 0 | 0 | | 0% |
| 7 | "Reconfiguring for Agility: Examining the Performance Implications of Project Team Autonomy through an Organizational Policy Experiment." | 1 | 1 | | | | | 100% |
| 8 | "Why do stores drive online sales? Evidence of underlying mechanisms from a multichannel retailer." | 0 | x | x | x | | | 0% |
| 9 | "Extrinsic versus intrinsic rewards for contributing reviews in an online platform." | 0 | 0 | 1 | x | 0 | | 25% |
| 10 | "Estimating the impact of "humanizing" customer service chatbots." | 1 | 1 | x | | | | 100% |
| 11 | "Impact of Customer Compensation Strategies on Outcomes and the Mediating Role of Justice Perceptions: A Longitudinal Study of Target's Data Breach." | 0 | 0 | x | 1 | 1 | 1 | 60% |
| 12 | "Effects of online recommendations on consumers' willingness to pay." | 1 | x | x | | | | 100% |
| 13 | "RECIPROCITY OR SELF-INTEREST? LEVERAGING DIGITAL SOCIAL CONNECTIONS FOR HEALTHY BEHAVIOR." | 0 | 0 | 0 | | | | 0% |
| 14 | "An empirical investigation of the antecedents and consequences of privacy uncertainty in the context of mobile apps." | 1 | 1 | x | 1 | 1 | | 100% |
| 15 | "Pictures that are worth a thousand donations: How emotions in project images drive the success of online charity fundraising campaigns? An image design perspective." | 1 | 1 | 1 | 1 | 1 | 1 | 100% |

*c1 means "conclusion 1," which is the first conclusion of a paper. The framework automatically identifies the ordinal of conclusions.

**Table 3: Participant Mode Results**

In participant mode, each session of the LLM is considered as a participant in the experiment. We prompt each variant of a field experiment five times to the LLM in separate sessions. For example, if the experiment has only one treatment and one control group, then each group will be prompted five times, as if there are five participants from the treatment group and five participants from the control group, totaling ten participants in the experiment. After we prompt them to LLM, we calculate the average treatment effect. If the average treatment effect's direction is the same as the direction of the actual conclusion, we mark it as 1, otherwise 0. It is worth mentioning that at the current stage, we can only simulate the direction of the treatment effect, not the size of the effect. For example, for a typical conclusion that training increases income by 15%, we are able to verify "increase" instead of "15%". Finally, we use "x" to represent that collected responses from the participant mode could not confirm or reject a conclusion. This may result from flaws in our participants' prompting generation or the design of the original paper. An example of the latter case is that a paper runs machine learning over field experimental results to get some conclusions. Simulated field experimental responses can neither confirm nor deny machine learning conclusions without running through the process.

We observe that participant mode results are notably less favorable compared to those of observer mode. We posit that the limitation does not stem from the LLM's inherent capability to simulate field experiments. In the automatic generation of participant mode promptings, converting the third-person perspective description of the workflow extracted from papers to the second-person perspective instructions for participants results in the loss of many details, leading to deviations from the real field experiment environment. Given these incomplete and altered experimental conditions, it is unsurprising that the LLM's responses diverge from those expected of actual human participants. Moreover, the process of generating questions for participant mode currently necessitates substantial human intervention, further complicating the simulation. Ongoing refinement of our methods is essential and will ultimately address these challenges, enhancing the accuracy and reliability of participant mode simulations.

## Data Memorization

| No. | Paper |
|-----|-------|
| 1 | "Friends with Health Benefits: A Field Experiment" (Gershon et al., 2024) Management Science |
| 2 | "Does Online Fundraising Increase Charitable Giving? A Nationwide Field Experiment on Facebook" (Adena & Hager, 2024) Management Science |
| 3 | "Improving Virtual Team Collaboration Paradox Management: A Field Experiment" (Luciano et al., 2024) Organization Science |
| 4 | "Put your mouth where your money is: A field experiment encouraging donors to share about charity" (Silver & Small, 2024) Marketing Science |
| 5 | "Uncovering Sophisticated Discrimination with the Help of Credence Goods Markups: Evidence from a Natural Field Experiment" (Hall et al., 2024) Management Science |

**Table 4: New Papers Published in 2024**

Data memorization is a common concern in simulating experiments with LLMs. If the results given by the LLM are from its memory of training data instead of reasoning, the proposed idea has no instructional value as pilot testing for field experiments (Horton, 2023). As revealed in its documents, the training data cutoff of GPT-4-turbo-2024-04-09 is December 2023. Therefore, we test observer and participant modes on five papers published in INFORMS in 2024, listed in Table 4. It is important to note that we check all five papers to make sure they or their preprint versions were not publicly available before December 2023. Therefore, it is less likely they were included in OpenAI's training data. While the results of the former 15 papers are 66% (observer) and 47.9% (participant), the results of the new five papers are 60.2% (observer) and 66.7% (participant), according to Table 5 and Table 6. By comparison, we show that the proposed approach works in new papers that are not likely to appear in the training data, likely mitigating the potential threat of data memorization to our research.

| No. | Paper | Observer Mode Results* | | | | | | |
|---|---|---|---|---|---|---|---|---|
| | | C1 | C2 | C3 | C4 | C5 | C6 | Avg (o) |
| 1 | "Friends with Health Benefits: A Field Experiment" | 75% | 93% | 96% | 96% | 96% | 96% | 90% |
| 2 | "Does Online Fundraising Increase Charitable Giving? A Nationwide Field Experiment on Facebook" | 0% | 0% | 0% | 13% | 93% | | 21% |
| 3 | "Improving Virtual Team Collaboration Paradox Management: A Field Experiment" | 0% | 100% | 100% | 93% | 93% | 93% | 80% |
| 4 | "Put your mouth where your money is: A field experiment encouraging donors to share about charity" | 0% | 25% | 18% | 4% | 4% | | 10% |
| 5 | "Uncovering Sophisticated Discrimination with the Help of Credence Goods Markups: Evidence from a Natural Field Experiment" | 100% | 100% | 100% | 100% | | | 100% |

*c1 means "conclusion 1," which is the first conclusion of a paper. The framework automatically identifies the ordinal of conclusions.

**Table 5: Observer Mode Results**

| No. | Paper | Participant Mode Results* | | | | | | |
|---|---|---|---|---|---|---|---|---|
| | | C1 | C2 | C3 | C4 | C5 | C6 | Avg (p) |
| 1 | "Friends with Health Benefits: A Field Experiment" | 1 | 1 | x | x | x | x | 100% |
| 2 | "Does Online Fundraising Increase Charitable Giving? A Nationwide Field Experiment on Facebook" | 1 | x | 1 | 1 | x | | 100% |
| 3 | "Improving Virtual Team Collaboration Paradox Management: A Field Experiment" | 1 | 1 | 1 | x | x | | 100% |
| 4 | "Put your mouth where your money is: A field experiment encouraging donors to share about charity" | 0 | 0 | 0 | x | x | | 0% |
| 5 | "Uncovering Sophisticated Discrimination with the Help of Credence Goods Markups: Evidence from a Natural Field Experiment" | 0 | x | 0 | 1 | | | 33% |

*c1 means "conclusion 1," which is the first conclusion of a paper. The framework automatically identifies the ordinal of conclusions.

**Table 6: Participant Mode Results**

## Discussion and Future Work

This study makes several significant contributions to the field. First, we expand the existing literature on LLMs' emergent capabilities by demonstrating, for the first time, that LLMs have the potential to simulate field experiments. To the best of our knowledge, this is the first work in literature that considers simulating field experiments that require a more complex environment setting and workflow design compared to the simulation of lab studies. Second, we introduce two novel prompting strategies: observer model, which directly predicts the main hypotheses based on key experimental settings, and participant model, which engages the LLM as a virtual participant in the experiment. Our third contribution lies in the fact that we can identify the potential boundaries within which LLMs work well for field experimental simulations. For example, we show that LLMs can correctly predict most conclusions of field experiments that are not based on gender differences or social norms. Finally, our framework enables researchers to conduct innovative field experimental simulations by configuring the necessary settings, as demonstrated in the 'optional' section of Figure 1.

Despite these contributions, our current approach has several limitations. One issue for observer mode is that, although we empirically conclude a few scenarios are infeasible for LLM-based simulation, the boundary between the feasible experimental scenarios and the infeasible is still unclear, impairing our approach's fidelity as pilot testing for field experiments. In participant mode, flaws in the automated prompt generation process may lead to incomplete or inaccurate simulated responses, hindering the verification of certain experimental conclusions. Additionally, the relatively low accuracy of participant mode diminishes the credibility of our proposed approach. This reduced accuracy may stem from the previously mentioned issues with prompt generation or inherent limitations in the LLM's ability to handle certain topics, such as those related to gender disparities.

## Reference

Aher, G. V., Arriaga, R. I., & Kalai, A. T. (2023, July). Using large language models to simulate multiple humans and replicate human subject studies. In *International Conference on Machine Learning* (pp. 337-371).

Charness, G., Jabarian, B., & List, J. A. (2023). *Generation next: Experimentation with ai* (No. w31679). National Bureau of Economic Research.

Dennis, A. R., & Valacich, J. S. (2001). Conducting experimental research in information systems. *Communications of the association for information systems*, *7*(1), 5.

Gneezy, A. (2017). Field experimentation in marketing research. *Journal of Marketing Research*, *54*(1), 140-143.

Harrison, G. W., & List, J. A. (2004). Field experiments. *Journal of Economic literature*, *42*(4), 1009-1055.

Horton, J. J. (2023). *Large language models as simulated economic agents: What can we learn from homo silicus?* (No. w31122). National Bureau of Economic Research.

Leibbrandt, A., & List, J. A. (2015). Do women avoid salary negotiations? Evidence from a large-scale natural field experiment. *Management Science*, *61*(9), 2016-2024.

Leng, Y., & Yuan, Y. (2023). Do LLM Agents Exhibit Social Behavior?. *arXiv preprint arXiv:2312.15198*.

Luo, X., Rechardt, A., Sun, G., Nejad, K. K., Yáñez, F., Yilmaz, B., ... & Love, B. C. (2024). Large language models surpass human experts in predicting neuroscience results. *arXiv preprint arXiv:2403.03230*.

Manning, B. S., Zhu, K., & Horton, J. J. (2024). Automated Social Science: A Structural Causal Model-Based Approach.

Yuan, Y., Tang, K., Shen, J., Zhang, M., & Wang, C. (2024). Measuring Social Norms of Large Language Models. *arXiv preprint arXiv:2404.02491*.

Zhao, J., Ding, Y., Jia, C., Wang, Y., & Qian, Z. (2024). Gender Bias in Large Language Models across Multiple Languages. *arXiv preprint arXiv:2403.00277*.